\documentclass[letterpaper]{article} 
\usepackage{aaai2026}  
\usepackage[hyphens]{url}  
\usepackage{graphicx} 
\usepackage{natbib}  
\usepackage{caption} 
\usepackage{amsmath}
\usepackage{booktabs}
\usepackage{array}
\usepackage{tabularx}
\usepackage{multirow}
\newif\ifshowauthors

\showauthorstrue      

\newcommand{\best}[1]{\textbf{#1}}
\newcolumntype{Y}{>{\raggedright\arraybackslash}X}

\title{Pointing-VLA: Typed Spatial Grounding Interfaces for Vision-Language-Action Manipulation}
\ifshowauthors
\author{
Xiwen Chen\textsuperscript{1,2,*},
Zelin Li\textsuperscript{1,*,\textdagger},
Zhiruo Zhou\textsuperscript{1,3},
Huiming Chen\textsuperscript{4}
Chenwei Wang\textsuperscript{5},
Xiaojun Zhu\textsuperscript{1,\textdagger}
}
\affiliations{
\textsuperscript{1}SIGS, Tsinghua University, China; \textsuperscript{2}Shanghai Jiao Tong University, China; 
\textsuperscript{3}Wuhan University of Technology, China;
\textsuperscript{4}City University of Hong Kong, Hong Kong (China SAR);
\textsuperscript{5}AiDlab, Hong Kong (China SAR);
\\
\textsuperscript{*}Equal Contributions; \textsuperscript{\textdagger}Correspondence: chiellini.lee@gmail.com; 
zhu.xiaojun@sz.tsinghua.edu.cn
}
\else
\author{Anonymous Submission}
\affiliations{}
\fi

\begin{document}

\maketitle

\raggedbottom

\begin{abstract}
Vision-language-action (VLA) models often expose spatial grounding through autoregressive text coordinates or opaque action tokens, creating brittle interfaces between multimodal reasoning and robot execution. We present Pointing-VLA, a typed hidden-state spatial readout built on Embodied-R1. Geometry-specific heads predict normalized points, object-functional grounding (OFG) heatmaps, and visual trajectories without serializing geometry as text. For the evaluated Bridge/WidowX and physical pick-place deployments, an explicit execution contract assigns PICK to source-conditioned OFG and PLACE to Pointing, providing direct stage-aligned spatial targets. Pointing-VLA achieves SOTA performance on Bridge/WidowX, averaging 72.9\% across the evaluated four-task set without Bridge-specific finetuning under collision-enabled CuRobo execution. Pointing and OFG show complementary strengths across native and cross-dataset evaluations. The OFG/contact readout transfers to NORA-1.5, preserving or improving success while reducing recorded controller time by more than 20$\times$; typed heads are also 6.68--6.90$\times$ faster than Embodied-R1 text decoding on a shared external suite. When integrated as spatial guidance for a $\pi_{0.5}$ action policy, Pointing-VLA raises autonomous real-robot success from 52.7\% to 80.7\% across three visual contexts. These results establish typed spatial readouts as an efficient, inspectable interface between embodied reasoning and robot execution.
\end{abstract}

\section{Introduction}

Large vision-language-action (VLA) models can connect observations, language instructions, and robot behavior, but their output interface remains a bottleneck. Many manipulation subtasks first require an explicit spatial commitment: the point to touch, the functional part to use, or the short visual trace an executor should follow. Autoregressive text coordinates are convenient for prompting, yet they make geometry an afterthought: the model must spend tokens spelling out numbers, the evaluator must parse strings back into pixels, and dense part-level evidence is discarded before execution. Direct action tokens avoid parsing, but they also hide the spatial rationale that a robot stack or human auditor may need to inspect.

\begin{figure*}[t]
\centering
\includegraphics[width=\textwidth]{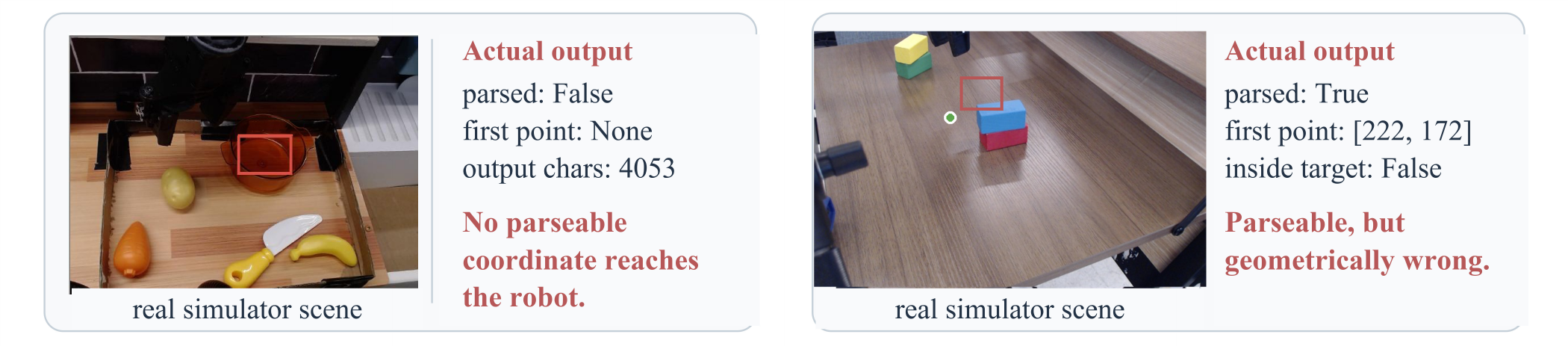}
\caption{Representative text-coordinate failures. \textbf{Left:} generation yields no parseable coordinate. \textbf{Right:} \texttt{[222, 172]} is parseable but lies outside the target region.}
\label{fig:text-coordinate-failures}
\end{figure*}

Figure~\ref{fig:text-coordinate-failures} separates two interface failures observed in representative simulator cases: text generation can fail before producing valid geometry, and a syntactically valid coordinate can still be spatially wrong. Parsing is therefore an additional execution dependency rather than a guarantee of correct grounding.

The central claim of this paper is that embodied grounding should be treated as typed interface prediction rather than text-coordinate generation. Referring points, functional regions, and visual trajectories are related but not identical targets: a point is sparse, an affordance is often a dense part-level region, and a trajectory is temporally structured. Collapsing them into one output distribution can create negative interference even when they share a visual-language backbone, while serializing all of them as text makes downstream execution slower and less stable.

This formulation is grounded in the vision-for-action and population-readout view of computation: downstream readouts decode spatial variables from distributed internal activity instead of serializing them through the system that represents them \cite{goodale1992,andersen2002,cisek2007}. Pointing-VLA realizes this principle with separate point, affordance-heatmap, and trajectory heads that emit each target in the geometry consumed by robot-side execution.

We propose Pointing-VLA, a lightweight typed grounding framework built on an embodied vision-language backbone. Rather than asking the backbone to say coordinates, Pointing-VLA reads spatial intent from multimodal hidden states. A compact adapter and learned queries feed point, object-functional grounding (OFG), and visual trajectory generation (VTG) decoders. Deployment contracts bind each structured execution slot to the geometry it requires; the primary pick-place scaffold uses source-conditioned OFG for PICK and Pointing for PLACE. The resulting spatial targets connect embodied reasoning to downstream wrappers, planners, and executors through deterministic, stage-aligned contracts.
Experiments evaluate the interface from native spatial grounding through head specialization, cross-backbone transfer, and robot deployment. On Bridge/WidowX, the fixed OFG PICK and Pointing PLACE composition achieves SOTA performance, averaging 72.9\% across the evaluated four-task set without Bridge-specific finetuning under collision-enabled CuRobo execution. Cross-dataset results reveal complementary Pointing and OFG geometries, while NORA-1.5 transfer preserves or improves success with more than 20$\times$ shorter recorded controller time. Real-robot deployment completes the evidence chain from hidden-state readout to physical execution.

Our contributions are:
\begin{itemize}
\item We formulate embodied grounding as typed interface prediction, replacing serialized text coordinates with geometry-specific robot-facing readouts.
\item We introduce Pointing, OFG, and VTG hidden-state readouts with explicit execution contracts, including a fixed source-conditioned OFG-PICK/Pointing-PLACE deployment that aligns each stage with its required geometry.
\item We validate the interface through native and cross-dataset grounding, runtime and cross-backbone transfer, Bridge/WidowX evaluation, and autonomous real-robot deployment.
\end{itemize}

\begin{figure*}[t]
\centering
\includegraphics[width=\textwidth]{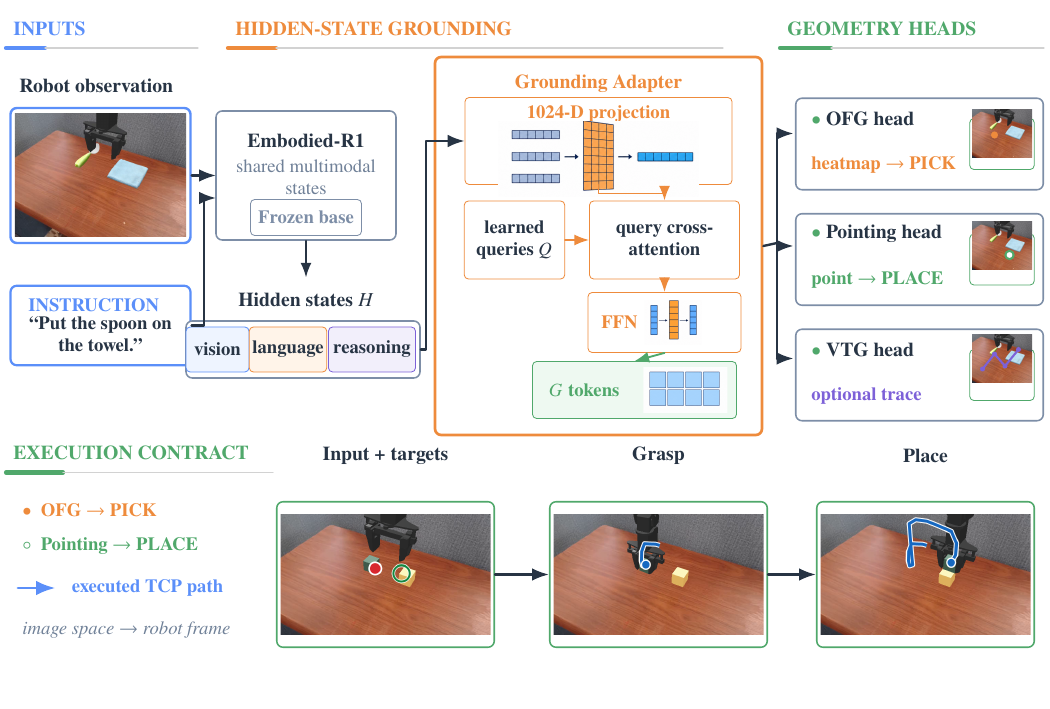}
\caption{Pointing-VLA architecture and typed interfaces. Embodied-R1 hidden states feed point, OFG/contact-heatmap, and VTG-waypoint heads; fixed PICK/PLACE slots use source-conditioned OFG/Pointing, and a wrapper maps image geometry to robot coordinates.}
\label{fig:architecture}
\end{figure*}

\begin{figure*}[t]
\centering
\includegraphics[width=\textwidth]{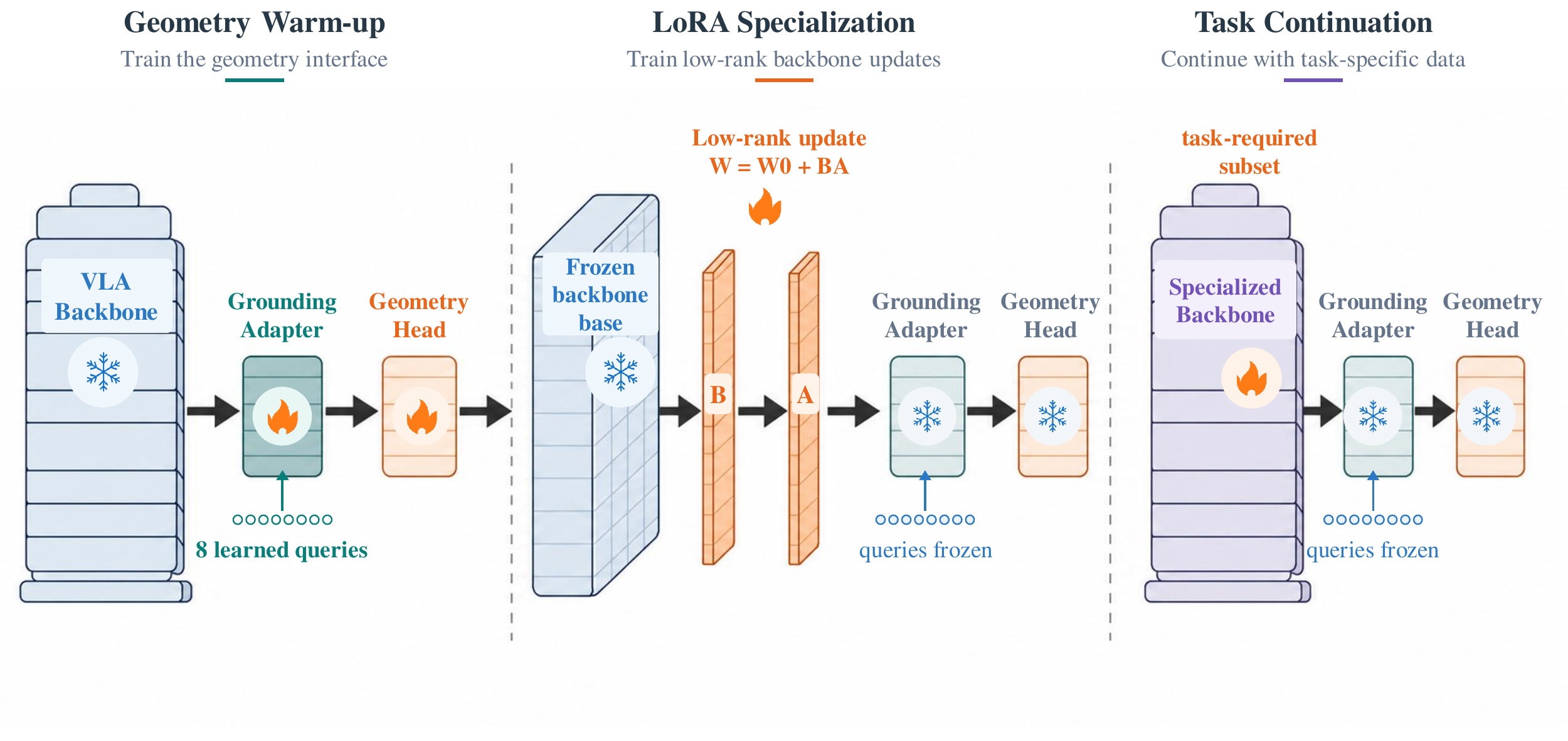}
\caption{Stage-wise training. Geometry warm-up trains the adapter, eight queries, and geometry head; LoRA specialization updates only low-rank backbone parameters; task continuation updates only the required subset. Snowflakes and flames mark frozen and trainable modules.}
\label{fig:training-schedule}
\end{figure*}

\section{Related Work}

Generalist robot policies and VLA systems adapt large pretrained models to robot action spaces, from embodied multimodal models and Robotics Transformers to cross-embodiment datasets and open robot policies \cite{palme2023,rt12023,rt22023,openx2024,octo2024,openvla2024,pizero2024}. This line of work has improved policy learning, action parameterization, and inference throughput, but it mainly asks how perception and language should become actions. Pointing-VLA asks a complementary interface question: what spatial representation should be exposed before final execution?

Spatial structure has long been useful for manipulation. Neuroscience distinguishes object recognition from action-facing visuomotor representations, including posterior-parietal intention maps and competing affordance-like action candidates \cite{goodale1992,andersen2002,cisek2007}. Robot learning makes a compatible engineering move: Transporter Networks preserve spatial feature maps for pick-and-place displacements, CLIPort combines semantic ``what'' and spatial ``where'' pathways, and PerAct discretizes 3D voxel actions for language-conditioned manipulation \cite{transporter2020,cliport2021,peract2023}. Recent grounding-centric VLA work extends this idea to embodied reasoning: RoboPoint predicts spatial affordance keypoints from language instructions \cite{robopoint2025}, FSD studies seeing-to-doing affordance grounding \cite{fsd2025}, and Embodied-R1 formalizes embodied pointing across REG, relational-region grounding, OFG, and VTG \cite{embodiedr12025}. Affordance datasets such as AGD20K further emphasize that actionable regions are often object parts rather than whole-object centers \cite{agd20k2022}. These systems show that spatial outputs can bridge VLM reasoning and robot behavior, but they leave open which interface an executor should consume. Pointing-VLA studies this interface choice directly: sparse referring points, dense functional heatmaps, and visual waypoint traces are decoded in their native geometries rather than serialized as text coordinates.

This interface complements action-sequence policies such as Diffusion Policy and ACT \cite{diffusionpolicy2023,act2023} by exposing typed spatial targets before low-level action generation. In the evaluated pick-place systems, the execution phase itself supplies a stable contract: functional contact is decoded for PICK and a compact target point is decoded for PLACE.

\section{Method}

\subsection{Hidden-State Spatial Readout}

Pointing-VLA operationalizes typed interface prediction by decoding geometry-specific robot targets directly from shared multimodal hidden states. Figure~\ref{fig:architecture} summarizes this hidden-state readout and its execution-facing spatial outputs.

Let $e\in\{\mathrm{pt},\mathrm{ofg},\mathrm{vtg}\}$ index a spatial expert and let $H_{\mathrm{mm}}^e=\{h_i^e\}_{i=1}^{N}$ denote its Embodied-R1 multimodal states. Pointing-VLA reuses one backbone architecture, base weights, and Grounding Adapter body, while each specialist retains its own LoRA, learned queries, and decoder state. For the point and OFG experts, the shared adapter body projects the sequence into a grounding space and reads it with expert-specific task-conditioned queries:
\[
\begin{split}
X_e &= W_h^e H_{\mathrm{mm}}^e + P_e,\\
A_e &= \operatorname{softmax}\!\left(
\frac{(Q_eW_Q^e)(X_eW_K^e)^\top}{\sqrt d}\right),\\
G_e &= \operatorname{FFN}_e\!\left(A_eX_eW_V^e\right),
\qquad g_e=\operatorname{Pool}(G_e),
\end{split}
\]
where $P_e$ supplies sequence position, $G_e$ preserves a fixed set of grounded tokens, and $g_e$ is its global summary. The point and OFG experts use eight grounding queries. The strongest VTG expert instead lets learned temporal queries attend directly to the full multimodal sequence, avoiding a fixed grounding-token bottleneck.

\subsection{Spatial Interfaces and Deployment Contracts}

\paragraph{Learned Spatial Heads.}
The Pointing decoder uses a learned pointer query $q_{\mathrm{pt}}$ to select the grounded evidence needed for one image location:
\[
\begin{split}
A_{\mathrm{pt}}&=\operatorname{softmax}\!\left(
\frac{(q_{\mathrm{pt}}W_Q)(G_{\mathrm{pt}}W_K)^\top}{\sqrt d}\right),\\
z_{\mathrm{pt}}&=A_{\mathrm{pt}}(G_{\mathrm{pt}}W_V),\qquad
(\hat x,\hat y)=\sigma\!\left(W_p\operatorname{LN}(z_{\mathrm{pt}})+b_p\right).
\end{split}
\]
Thus $(\hat x,\hat y)\in[0,1]^2$ is emitted directly by the learned head, without generating or parsing coordinate text. For an image of width $W$ and height $H$, the corresponding pixel is
\[
(\hat u,\hat v)=\bigl((W-1)\hat x,(H-1)\hat y\bigr).
\]
The prediction is evaluated with point-in-bbox (PIB) or point-in-mask (PIM).

OFG preserves spatial extent instead of collapsing the source to one token. Let $F_{\mathrm{vis}}$ be the visual tower's 2D feature map. The global grounded summary modulates this map through FiLM before a convolutional heatmap decoder \cite{agd20k2022}:
\[
\begin{split}
F'_{\mathrm{ofg}}&=\gamma(g_{\mathrm{ofg}})\odot F_{\mathrm{vis}}
+\beta(g_{\mathrm{ofg}}),\\
\hat H_{\mathrm{ofg}}&=f_{\mathrm{hm}}(F'_{\mathrm{ofg}}),\\
(i^*,j^*)&=\arg\max_{i,j}\hat H_{\mathrm{ofg}}[i,j],\\
\hat p_{\mathrm{ofg}}&=\left(\frac{j^*}{W_o-1},\frac{i^*}{H_o-1}\right).
\end{split}
\]
Here $\hat H_{\mathrm{ofg}}$ contains heatmap logits, and its normalized peak is the execution-facing functional contact.

For VTG, learned waypoint queries $q_t^\tau$ with temporal encodings attend to the multimodal sequence in order:
\[
\begin{split}
z_t^\tau&=\operatorname{Attn}(q_t^\tau,H_{\mathrm{mm}}^{\mathrm{vtg}}),\\
\hat\tau_t&=\sigma(W_\tau z_t^\tau+b_\tau),\qquad t=1,\ldots,T.
\end{split}
\]
We use $T=8$ normalized image-space waypoints and report RMSE, average displacement error (ADE), and final displacement error (FDE). These waypoints describe a visual trace rather than low-level robot actions.

\paragraph{Structured PICK/PLACE Contracts.}
Training follows target geometry: point, region, and visual-trace samples supervise the corresponding learned heads. For pick-place deployment, the scaffold constructs slot-specific queries from instruction $\ell$, source description $c_{\mathrm{src}}$, and goal description $c_{\mathrm{goal}}$:
\[
\begin{array}{rcl}
q_{\mathrm{pick}}&=&\mathcal{T}_{\mathrm{pick}}(\ell,c_{\mathrm{src}}),\\
q_{\mathrm{place}}&=&\mathcal{T}_{\mathrm{place}}(\ell,c_{\mathrm{goal}}),\\
\hat y_{\mathrm{pick}}&=&F_{\mathrm{ofg}}(H_{\mathrm{mm}},G,q_{\mathrm{pick}}),\\
\hat y_{\mathrm{place}}&=&F_{\mathrm{pt}}(H_{\mathrm{mm}},G,q_{\mathrm{place}}).
\end{array}
\]
This assignment is deterministic: source-conditioned OFG emits the functional PICK contact, while Pointing emits the PLACE target. VTG serves annotated waypoint requests; the reported pick-place deployments use the fixed OFG-PICK/Pointing-PLACE contract. An external wrapper converts image-space outputs to robot-frame targets. With calibrated depth $D$, camera intrinsics $K$, and camera-to-robot transform $T_{r\leftarrow c}$, this boundary can be written as
\[
\begin{split}
\tilde p&=[\hat u,\hat v,1]^\top,\\
X_c&=D(\hat u,\hat v)K^{-1}\tilde p,\\
\bar X_r&=T_{r\leftarrow c}\bar X_c,
\end{split}
\]
where $\bar X$ denotes homogeneous coordinates. The existing executor then consumes $X_r$; the learned heads remain image-space predictors.

\subsection{Training Objective}

The experts use a weighted spatial objective:
\[
\mathcal{L} =
\lambda_{\mathrm{pt}}\mathcal{L}_{\mathrm{pt}}
+ \lambda_{\mathrm{box}}\mathcal{L}_{\mathrm{box}}
+ \lambda_{\mathrm{ofg}}\mathcal{L}_{\mathrm{ofg}}
+ \lambda_{\mathrm{vtg}}\mathcal{L}_{\mathrm{vtg}}
+ \lambda_{\mathrm{aux}}\mathcal{L}_{\mathrm{aux}}.
\]
Point and trajectory coordinates use SmoothL1, while OFG mask or keypoint supervision is converted to a dense target $H^*$:
\[
\begin{array}{rcl}
\mathcal{L}_{\mathrm{pt}}&=&\frac{1}{K}\sum_k\mathrm{SmoothL1}(\hat p_k,p_k),\\
\mathcal{L}_{\mathrm{ofg}}&=&\frac{1}{|\Omega|}\sum_{u\in\Omega}
\ell_{\mathrm{hm}}(\hat H_{\mathrm{ofg}}(u),H^*(u)),\\
\mathcal{L}_{\mathrm{vtg}}&=&\frac{1}{T}\sum_t\mathrm{SmoothL1}(\hat\tau_t,\tau_t).
\end{array}
\]
Here $\ell_{\mathrm{hm}}$ is mean-squared error for Gaussian keypoint targets and class-balanced binary cross-entropy with logits for dense affordance masks. Language-modeling and legacy reconstruction terms are inactive; optimization focuses on adapters, heads, and LoRA specialization \cite{lora2021}.

For source-conditioned OFG specialization, we keep the Embodied-R1 base, Pointing/VTG query and decoder modules, and executor frozen while jointly updating the OFG LoRA, shared Grounding Adapter body, OFG learned query, and heatmap decoder. Dense source-mask reconstruction is combined with source-over-distractor ranking and distractor suppression. This adaptation teaches OFG to preserve functional contact while resolving the source instance directly in the PICK readout.

The final pointing model additionally applies GRPO-style group-normalized refinement. With policy center $\mu=\hat p$ and learned $\sigma=\exp(s)$, samples are
\[
a_i=\mathrm{clip}(\mu+\sigma\epsilon_i,0,1),\qquad
\epsilon_i\sim\mathcal{N}(0,I).
\]
For target box $B^*$, reward and group-normalized advantage are
\[
\begin{array}{rcl}
r_i&=&\left\{\begin{array}{ll}
1,&a_i\in B^*,\\
0.5\max(0,1-d_i/\rho),&a_i\notin B^*,
\end{array}\right.\\[2pt]
A_i&=&\frac{r_i-\mathrm{mean}_j(r_j)}{\mathrm{std}_j(r_j)+\epsilon}.
\end{array}
\]
With $\pi_i=\pi(a_i\mid\mu,\sigma)$ and Gaussian reference KL $D_{\mathrm{ref}}$, the objective is
\[
\begin{array}{rcl}
\mathcal{L}_{\mathrm{GRPO}}
&=& -\frac{1}{M}\sum_{i=1}^{M}\mathrm{sg}(A_i)\log \pi_i\\
&& + \lambda_{\mathrm{sft}}\mathrm{SmoothL1}(\mu,p^*)\\
&& + \lambda_{\mathrm{kl}}D_{\mathrm{ref}}
 + \lambda_{\sigma}\mathcal{L}_{\sigma}.
\end{array}
\]
Here $\mathrm{sg}$ stops advantage gradients and $p^*$ is the supervised point target; refinement acts directly on the continuous spatial head.

\section{Experiments}

\subsection{Evaluation Questions and Protocol}

Four questions structure the evaluation.
\par\noindent\textbf{Q1: Geometry.} Do typed heads preserve task-appropriate geometry?
\par\noindent\textbf{Q2: Phase specialization.} Does source-conditioned OFG resolve PICK source selection while preserving the benefits of a fixed OFG-PICK/Pointing-PLACE contract?
\par\noindent\textbf{Q3: Transfer and efficiency.} Does the readout transfer across datasets and VLA backbones while reducing inference cost?
\par\noindent\textbf{Q4: Deployment.} Do the gains persist in simulation and real-robot deployment?

Native and external evaluations answer Q1; source-conditioning and interface-composition ablations answer Q2. These controlled comparisons isolate component-level grounding quality, while the deployment studies measure complete robot-task success. Cross-dataset and cross-backbone transfer plus known-head runtime address Q3; controlled simulation and real-robot studies on shared task sets and executor families address Q4.

\paragraph{Evaluation Protocol and Metrics.}
We evaluate each output at the level of the interface it is designed to serve. Pointing and OFG predictions are converted to image coordinates and scored by point-in-mask (PIM) or point-in-box (PIB) sample success; VTG is measured by normalized trajectory root mean squared error (RMSE), average displacement error (ADE), and final displacement error (FDE). Robot studies require full task completion. Within each experiment, all compared methods use the same samples or episode identities, ensuring that measured differences arise from the spatial interface or adaptation strategy. Experiments used 40-GB NVIDIA A100 and dual 48-GB RTX A6000 GPUs; complete hardware, software, and seed records appear in the Supplement.

\subsection{Native Spatial-Head Results}

Table~\ref{tab:native} answers the first question. Pointing-VLA matches Embodied-R1 on sparse referring-point grounding at 64.3\%, while OFG raises full Part-Affordance-2K accuracy from 40.9\% to 57.3\%. This 16.4-point improvement demonstrates geometry specialization: Pointing preserves sparse target accuracy, whereas OFG substantially strengthens part-level functional grounding by retaining the spatial extent required for contact prediction.

\begin{table}[!htbp]
\centering
\footnotesize
\setlength{\tabcolsep}{2.5pt}
\begin{tabular*}{\columnwidth}{@{\extracolsep{\fill}}lcc@{}}
\toprule
Model & \shortstack{VABench-P\\acc. $\uparrow$} & \shortstack{Part-Afford.2K\\acc. $\uparrow$} \\
\midrule
GPT-4o & 9.3 & 10.2 \\
ASMv2 & 10.1 & 13.8 \\
RoboBrain & 7.0 & 25.3 \\
Qwen2.5-VL & 9.9 & 23.4 \\
RoboPoint \cite{robopoint2025} & 19.1 & 27.6 \\
FSD \cite{fsd2025} & 61.8 & 9.6 \\
Embodied-SFT \cite{embodiedr12025} & 50.5 & 39.4 \\
Embodied-R1 \cite{embodiedr12025} & 63.7 & 40.9 \\
Pointing-VLA (ours) & \best{64.3} & \best{57.3} \\
\bottomrule
\end{tabular*}
\caption{Native point-in-region grounding accuracy (\%). Pointing-VLA uses its Pointing head for VABench-P and its OFG head for the full Part-Affordance-2K run.}
\label{tab:native}
\end{table}

\textbf{Sequential geometry.} VTG completes the native evaluation for ordered spatial outputs. On 300 VABench-V examples, the trajectory readout records 0.1042 RMSE, 0.1368 ADE, and 0.1493 FDE in normalized image coordinates. These measures evaluate the waypoint sequence and its terminal target rather than collapsing trajectory quality into a single point-in-region score.

\subsection{Mechanism and Fixed-Contract Ablations}

PICK requires both source-instance disambiguation and a contact location within the selected object. Source-conditioned OFG preserves this dense functional extent and exposes its peak as the grasp cue. PLACE instead requires a compact target in the destination region, for which Pointing provides a direct normalized coordinate. The fixed OFG-PICK/Pointing-PLACE contract therefore aligns each execution stage with the geometry it requires, without introducing an inference-time selector.

\begin{table*}[!t]
\centering
\small
\begin{minipage}[t]{0.35\textwidth}
\centering
\textbf{(a) Spatial-head configuration}\\[2pt]
\setlength{\tabcolsep}{2.0pt}
\begin{tabularx}{\linewidth}{@{}lYc@{}}
\toprule
Component & Settings & PIB \\
\midrule
Query tokens & 4 / 8 / 16 & 44.7 / \best{50.3} / 45.7 \\
LoRA rank & none / 16 / 32 / 64 & 45.3 / 47.3 / \best{53.3} / 50.3 \\
Refinement & SFT / GRPO & 61.3 / \best{64.3} \\
\bottomrule
\end{tabularx}
\end{minipage}
\hfill
\begin{minipage}[t]{0.63\textwidth}
\centering
\textbf{(b) Interface-composition ablation}\\[2pt]
\setlength{\tabcolsep}{1.6pt}
\begin{tabular*}{\linewidth}{@{\extracolsep{\fill}}lccccc@{}}
\toprule
Configuration & \shortstack{Spoon\\on towel} & \shortstack{Carrot\\on plate} & \shortstack{Stack\\cube} & \shortstack{Eggplant\\basket} & Avg. \\
\midrule
Attention $\rightarrow$ Attention & 16.7 & 41.7 & 100.0 & 62.5 & 55.2 \\
Attention $\rightarrow$ Pointing & 12.5 & 50.0 & 100.0 & 70.8 & 58.3 \\
OFG $\rightarrow$ Attention & 16.7 & 54.2 & 16.7 & 58.3 & 36.5 \\
\textbf{OFG $\rightarrow$ Pointing} & \best{50.0} & \best{50.0} & \best{100.0} & \best{91.7} & \best{72.9} \\
\bottomrule
\end{tabular*}
\end{minipage}
\caption{Mechanism and fixed-contract results. (a) VABench-P PIB (\%) across spatial-head configurations. (b) Bridge/WidowX success (\%) for fixed PICK/PLACE interfaces (24 episodes/task); the final row is the deployed OFG-PICK/Pointing-PLACE contract with collision-enabled CuRobo.}
\label{tab:pointing-ablation-panels}
\label{tab:phase-aligned-composition-ablation}
\end{table*}

\textbf{Mechanism analysis.} Panel (a) identifies the effective capacity regime of the spatial readout. Eight queries and rank-32 LoRA produce the strongest PIB within their respective sweeps, while GRPO refinement raises PIB from 61.3\% to 64.3\%. The non-monotonic query and rank trends demonstrate that performance is governed by the spatial-readout design rather than parameter count alone.

\textbf{Fixed-contract validation.} Panel (b) validates the phase-aligned assignment: the deployed contract completes 70/96 episodes (72.9\%) under collision-enabled CuRobo execution, outperforming the strongest alternative composition by 14.6 percentage points while completing all Stack episodes and 22/24 Eggplant episodes.

The controlled frozen multicolor Stack evaluation further verifies this mechanism. Source-conditioned OFG selects the instructed source in all 48 online cases and completes 43/48 episodes, versus 40/48 for Attention PICK under identical Pointing PLACE targets. Perfect source selection establishes OFG as a direct PICK readout that resolves the instructed source while preserving functional contact grounding.

\begin{figure*}[!t]
\centering
\includegraphics[width=\textwidth]{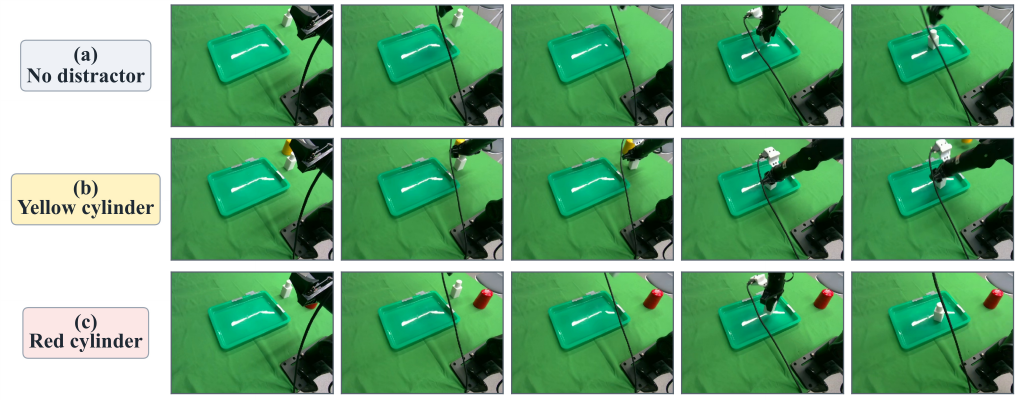}
\caption{Representative successful real-robot rollouts across three visual contexts, each ordered from initialization to stable release.}
\label{fig:real-robot-deployment}
\end{figure*}

\subsection{External Diagnostics and Cross-Backbone Transfer}

\paragraph{Cross-Dataset Geometry.}
Table~\ref{tab:external} asks whether performance follows the geometry exposed by each interface. These diagnostics standardize outputs as points and score PIM/PIB, isolating spatial grounding from downstream control. Pointing is strongest on RefCOCO and RoboAff, whereas OFG is strongest on AGD20K and raises full Part-Affordance-2K PIM from 40.9\% with Embodied-R1 text generation to 57.3\%. The crossover supports expert-specific output geometries rather than one serialized coordinate interface. The Supplement provides qualitative OFG heatmaps.

Pointing, OFG, and Embodied-R1 text generation are evaluated on the same cross-dataset samples. Each trained head emits one point, while text generation can emit multiple points; sample-level success therefore keeps the comparison independent of candidate count. Dataset composition and deterministic split construction are detailed in the Supplement and released evaluation code. This shared point protocol enables a controlled cross-dataset comparison of spatial interfaces under identical output and evaluation rules.

\begin{center}
\begin{minipage}{\columnwidth}
\centering
\scriptsize
\setlength{\tabcolsep}{1.8pt}
\renewcommand{\arraystretch}{0.92}
\begin{tabular*}{\linewidth}{@{\extracolsep{\fill}}llccc@{}}
\toprule
Suite & Metric & \shortstack{Pointing-VLA\\Point} & \shortstack{Pointing-VLA\\OFG} & \shortstack{Embodied-R1\\Text Generation} \\
\midrule
RefCOCO & PIM & \best{40.7\%} & 22.2\% & 10.6\% \\
RefCOCO & PIB & \best{56.1\%} & 36.0\% & 25.9\% \\
AGD20K & PIM & 53.9\% & \best{64.4\%} & 39.9\% \\
Part-Aff.2K & PIM & -- & \best{57.3\%} & 40.9\% \\
RoboAff. & PIM & \best{11.5\%} & 9.8\% & 2.4\% \\
\bottomrule
\end{tabular*}
\captionof{table}{Cross-dataset PIM/PIB accuracy (\%).}
\label{tab:external}
\end{minipage}
\end{center}

\begin{center}
\begin{minipage}{\columnwidth}
\centering
\scriptsize
\setlength{\tabcolsep}{1.6pt}
\renewcommand{\arraystretch}{0.92}
\begin{tabular*}{\linewidth}{@{\extracolsep{\fill}}lrrr@{}}
\toprule
System & Total (s) & Per sample (s) & Speedup \\
\midrule
OFG head & 1,274.61 & 0.434 & 6.90$\times$ \\
Pointing head & 1,316.29 & 0.448 & 6.68$\times$ \\
Embodied-R1 text decoder & 8,798.06 & 2.993 & 1.00$\times$ \\
\bottomrule
\end{tabular*}
\par\vspace{2pt}
\begin{tabular*}{\linewidth}{@{\extracolsep{\fill}}llrrr@{}}
\toprule
Method & Pose & Success & Time (s) & Speedup \\
\midrule
NORA base & Horizontal & 100.0 & 107.11 & 1.0$\times$ \\
+ OFG/contact & Horizontal & \best{100.0} & \best{4.85} & \best{22.1$\times$} \\
NORA base & Laid vert. & 89.0 & 111.57 & 1.0$\times$ \\
+ OFG/contact & Laid vert. & \best{95.0} & \best{5.44} & \best{20.5$\times$} \\
\bottomrule
\end{tabular*}
\captionof{table}{Known-head runtime (top) and NORA-1.5 transfer (bottom).}
\label{tab:runtime}
\label{tab:nora-transfer}
\end{minipage}
\end{center}

\paragraph{Interface-Level Runtime.}
Table~\ref{tab:runtime} measures inference from multimodal hidden states to typed spatial outputs under the shared external protocol. The geometric readouts are 6.68--6.90$\times$ faster than autoregressive text decoding, establishing the interface-level efficiency gained by eliminating coordinate serialization, token-by-token generation, and post-hoc parsing.

\paragraph{NORA-1.5 Transfer.}
Table~\ref{tab:nora-transfer} shows that the frozen NORA-1.5 backbone with the transferred OFG/contact readout preserves horizontal-pose success, improves laid-vertical success from 89.0\% to 95.0\%, and cuts recorded controller time by more than 20$\times$ under the shared wrapper.

\begin{center}
\begin{minipage}{\columnwidth}
\centering
\scriptsize
\setlength{\tabcolsep}{1.2pt}
\begin{tabular*}{\columnwidth}{@{\extracolsep{\fill}}lrrrrr@{}}
\toprule
System & Spoon & Carrot & Stack & Eggpl. & Avg. \\
\midrule
Octo-S & 47.2 & 9.7 & 4.2 & 56.9 & 30.0 \\
SpatialVLA (FT) & 16.7 & 25.0 & 29.2 & \best{100.0} & 42.7 \\
SoFar & 55.5 & 56.9 & 62.5 & 40.2 & 53.8 \\
MemoryVLA & 75.0 & 75.0 & 37.5 & \best{100.0} & 71.9 \\
Embodied-R1 + CuRobo & 20.8 & 45.8 & 62.5 & 79.2 & 52.1 \\
\textbf{Pointing-VLA + CuRobo} & 50.0 & 50.0 & \best{100.0} & 91.7 & \best{72.9} \\
\bottomrule
\end{tabular*}
\captionof{table}{Bridge/WidowX success (\%; 24 episodes/task for Embodied-R1 and Pointing-VLA). Pointing-VLA uses fixed OFG-PICK/Pointing-PLACE with collision-enabled CuRobo.}
\label{tab:simulation}
\end{minipage}
\end{center}

\subsection{Deployment Studies}
\label{sec:local-deploy}

Table~\ref{tab:simulation} evaluates four Bridge/WidowX manipulation structures. The Pointing-VLA condition contains 24 episodes per task, and success requires complete task execution. Its fixed OFG PICK and Pointing PLACE composition reaches 72.9\% with collision checking active in every rollout; CuRobo completes all planned motion segments without planner failure or fallback. The tasks stress distinct spatial bottlenecks: thin-object contact for Spoon, surface targeting for Carrot, source-instance disambiguation for Stack, and container placement for Eggplant. The source-conditioned OFG expert preserves 100.0\% Stack success as the direct PICK readout, while Pointing supplies an explicit placement target. Together, these results demonstrate reliable four-task deployment through a collision-aware motion-planning backend.

\subsection{Real-Robot Deployment}
\label{sec:real-robot-deployment}

\begin{center}
\begin{minipage}{\columnwidth}
\centering
\small
\setlength{\tabcolsep}{2.0pt}
\begin{tabular*}{\columnwidth}{@{\extracolsep{\fill}}lccc@{}}
\toprule
Scene success & $\pi_{0.5}$ & $\pi_{0.5}$ + P-VLA & $\Delta$ \\
\midrule
No distractor & 20/50 (40.0) & \best{36/50 (72.0)} & +32.0 \\
Yellow cylinder & 26/50 (52.0) & \best{42/50 (84.0)} & +32.0 \\
Red cylinder & 33/50 (66.0) & \best{43/50 (86.0)} & +20.0 \\
\midrule
All three scenes & 79/150 (52.7) & \best{121/150 (80.7)} & \best{+28.0} \\
\bottomrule
\end{tabular*}
\vspace{2pt}
\begin{tabular*}{\columnwidth}{@{\extracolsep{\fill}}lcccc@{}}
\toprule
Failure stage & Grasp & Transfer & Tray & Uncertain \\
\midrule
Count (/150) & $47\!\rightarrow\!16$ & $8\!\rightarrow\!9$ & $13\!\rightarrow\!4$ & $3\!\rightarrow\!0$ \\
$\Delta$ (pp) & \best{-20.7} & +0.7 & -6.0 & -2.0 \\
\bottomrule
\end{tabular*}
\captionof{table}{Autonomous PiPER outcomes. Success is count/trials (\%); failures show baseline$\rightarrow$P-VLA for pre-lift grasp, post-grasp transfer/drop, tray arrival without upright release, and uncertain cases.}
\label{tab:real-robot-deployment}
\end{minipage}
\end{center}

We deploy Pointing-VLA on an AgileX PiPER and compare it with a $\pi_{0.5}$ baseline \cite{intelligence2025pi05} through the same dual-camera perception-to-control stack. Pointing-VLA retains $\pi_{0.5}$ as the action-generating policy and inserts typed spatial guidance before execution. During PICK, the OFG head predicts an affordance heatmap over the lower rectangular part of the white composite object, and the external wrapper converts its peak into a robot-frame grasp cue. During PLACE, the Pointing head predicts a normalized image-space placement coordinate inside the green tray, which the wrapper converts into the placement target. The same deterministic OFG-PICK/Pointing-PLACE contract is used throughout all three visual contexts: no distractor, a yellow-cylinder distractor, and a red-cylinder distractor. We evaluate both systems over 50 trials per context. The reported endpoint is full task completion: the designated part must be grasped and lifted, positioned upright in the tray, and stably released.

Across all three visual contexts, typed spatial guidance consistently improves autonomous completion over the action-policy baseline. Failure-stage analysis confirms that the gains concentrate at the intended control boundaries: pre-lift grasp failures fall from 47 to 16, and tray-arrival failures fall from 13 to 4. Post-grasp transfer remains the primary residual bottleneck. These reductions verify the stage-aligned contribution of OFG-PICK and Pointing-PLACE guidance; Figure~\ref{fig:real-robot-deployment} traces representative successes from approach through stable release.

\section{Discussion and Conclusion}

Pointing-VLA treats referring points, functional regions, and waypoint traces as distinct robot-facing geometries. Embodied-R1 provides shared multimodal states, while lightweight geometry heads expose normalized points, heatmaps, and trajectories directly. Explicit deployment contracts align these outputs with execution stages and preserve the geometry required by each robot-facing slot. Together, the native, cross-dataset, runtime, transfer, and deployment results establish geometry-appropriate typed readouts as an efficient and inspectable interface between embodied VLM reasoning and robot execution.

The fixed pick/place scaffold provides a controlled deployment setting that verifies the effectiveness of typed spatial guidance. Task-dependent variation, most visible on spoon-on-towel, identifies closed-loop geometry-aware execution as the next leverage point. Future work will extend this interface with visual correction, broader structured task decompositions, and cross-robot transfer.

\bibliography{aaai2027}

\end{document}